\documentclass[10pt, a4paper]{article}
\usepackage{lrec2022} % this is the new LREC2022 Style
\pdfoutput=1
\usepackage{multibib}
\usepackage{graphicx}
\usepackage[frozencache,cachedir=mint]{minted} 
\usepackage{tabularx}
\usepackage{soul}
\usepackage{times}
\usepackage{listings}
\usepackage{caption}
\usepackage{stfloats}
\usepackage{latexsym}
\usepackage{multirow}
\usepackage{booktabs}
\usepackage{enumitem}
\usepackage{xcolor}
\usepackage{amsmath}

\DeclareMathOperator*{\argmin}{arg\,min}

\usepackage{titlesec}
\titleformat{\section}{\normalfont\large\bfseries\center}{\thesection.}{1em}{}
\titleformat{\subsection}{\normalfont\SmallTitleFont\bfseries\raggedright}{\thesubsection.}{1em}{}
\titleformat{\subsubsection}{\normalfont\normalsize\bfseries\raggedright}{\thesubsubsection.}{1em}{}
\renewcommand\thesection{\arabic{section}}
\renewcommand\thesubsection{\thesection.\arabic{subsection}}
\renewcommand\thesubsubsection{\thesubsection.\arabic{subsubsection}}
\usepackage{epstopdf}
\usepackage[utf8]{inputenc}

\usepackage{hyperref}
\usepackage{xstring}

\usepackage{color}

\title{Bootstrapping Text Anonymization Models with Distant Supervision}

\name{Anthi Papadopoulou$^1$, Pierre Lison$^2$, Lilja Øvrelid$^1$, Ildikó Pilán$^2$} 

\address{$^1$Language Technology Group, University of Oslo \\
            $^2$Norwegian Computing Center\\
         Oslo, Norway \\
         \begin{small}\textsf{\{anthip,liljao\}@ifi.uio.no}\end{small} \ \ \ \ \ \ \ \begin{small}\textsf{\{plison,pilan\}@nr.no}\end{small} \\}

\abstract{
We propose a novel method to bootstrap text anonymization models based on distant supervision.  Instead of requiring manually labeled training data, the approach relies on a knowledge graph expressing the background information assumed to be publicly available about various individuals. This knowledge graph is employed to automatically annotate text documents including personal data about a subset of those individuals. More precisely, the method determines which text spans ought to be masked in order to guarantee $k$-anonymity, assuming an adversary with access to both the text documents and the background information expressed in the knowledge graph. The resulting collection of labeled documents is then used as training data to fine-tune a pre-trained language model for text anonymization. We illustrate this approach using a knowledge graph extracted from Wikidata and short biographical texts from Wikipedia. Evaluation results with a RoBERTa-based model and a manually annotated collection of 553 summaries showcase the potential of the approach, but also unveil a number of issues that may arise if the knowledge graph is noisy or incomplete.  The results also illustrate that, contrary to most sequence labeling problems, the text anonymization task may admit several alternative solutions. 
 \\ \newline  }

\begin{document}

\maketitleabstract

\section{Introduction}

Personal data is ubiquitous in text documents. Due to this presence of personal information, many text sources fall under the scope of data protection regulations such as the General Data Protection Regulation (GDPR) recently introduced in Europe \cite{gdpr}. As a consequence, they cannot be shared with third parties (or even used for other purposes than the one originally intended when collecting the data) without a proper legal ground, such as the explicit consent of the individuals. 

In case obtaining the consent of all those individuals is unfeasible (for instance because there is not practical way of contacting the individuals in question), an alternative is to \textit{anonymize} the data to ensure those individuals can no longer be identified. Anonymization is often defined as the complete and irreversible process of removing all Personally Identifiable Information (PII) from a dataset \cite{soton399692}. Such PII includes both direct identifiers such as person names or passport numbers, but also more indirect information such as date of birth, gender or nationality that can also lead to \mbox{(re-)identification} when combined with one another \cite{Models}.

The anonymization of text data is, however, a difficult challenge for which many open questions remain \cite{acl2021}. One important problem is the lack of labeled corpora for this task, making it difficult to train data-driven text anonymization models in many domains. Setting up an anonymization task is costly and it requires good annotation guidelines and people at least familiar with data protection. The few datasets that currently exist mainly focus on the medical domain \cite{dernoncourt2017identification,brathen-etal-2021-creating} and are typically limited to predefined categories of entities\footnote{This task of detecting and masking predefined semantic categories (such as names, organizations and locations) is often called \textit{de-identification}. In contrast, text \textit{anonymization} is not limited to a fixed set of semantic categories, but must consider how any textual element may influence the risk of disclosing the identity of the person referred to in the text \cite{chevrier2019use,acl2021}.}. Models trained on such datasets are also known to be difficult to transfer to new domains \cite{johnson2020,hartman2020}.

\begin{figure*}[!t]
\includegraphics[width = 0.94\textwidth]{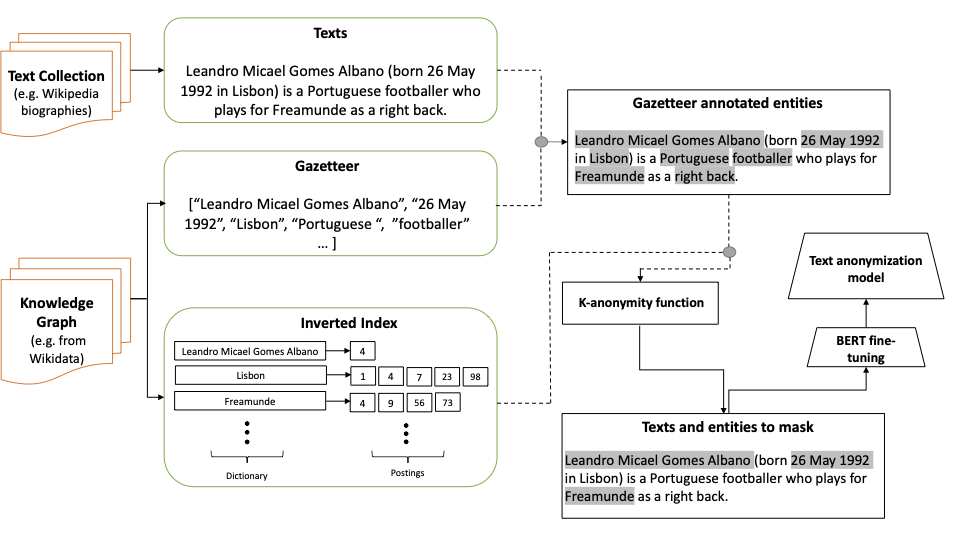}
    \caption{General sketch of the approach, illustrated with some examples for clarity}
     \label{fig:intro}
\end{figure*}

We present in this paper an alternative approach for training text anonymization models. Crucially, this approach does not require access to manually labeled training data. Rather, we adopt a distant supervision approach that revolves around a \emph{knowledge graph} expressing the background information assumed to publicly known on various individuals. The approach proceeds in three steps:
\begin{enumerate}
\itemsep0em
    \item The knowledge graph is first converted into an inverted index, making it possible to efficiently compute the set of individuals associated with a given combination of entities.
    \item The inverted index is then employed as distant supervision source \cite{mintz-etal-2009-distant,liang2020bond} to automatically annotate a collection of text documents including personal data. The goal of this annotation process is to determine which tokens to mask in order to guarantee $k$-anonymity (that is, to guarantee that the information conveyed in the anonymized document is sufficiently general to be shared by at least $k$ individuals).
    \item Finally, this labeled collection of documents is used as training data to fine-tune a large, pre-trained language model (in our case RoBERTa) for the task of determining which text span to mask in a given document.
\end{enumerate}

The proposed approach has several benefits. As it relies on distant supervision, there is no need for manually annotating text documents with text spans to mask, a procedure that is costly and time-consuming. The approach also follows a \textit{privacy-first} strategy that determines which terms to mask based on a privacy model ($k$-anonymity). This strategy provides an explicit account of the disclosure risk associated with a given set of masking decisions on a document, using the knowledge graph to represent the information that can be drawn upon by an adversary to uncover the identity of the individual(s) we seek to protect. This account of disclosure risk makes it possible to adjust the trade-off between data protection and data utility. Finally, the approach makes it arguably easier to port text anonymization models to new languages and domains, as knowledge graphs can often be reused across multiple languages and text genres.

The validity of the approach is evaluated through experiments with a collection of short biographical texts extracted from Wikipedia. Wikipedia biographies constitute an ideal test-bed for the proposed approach, as these texts contain a lot of PII, including both direct and quasi identifiers. Those biographical texts were then automatically annotated with text spans to masks using a knowledge graph derived from Wikidata. The general procedure is illustrated in Figure \ref{fig:intro}.

This paper makes three main contributions:
\raggedbottom
\begin{itemize}
\itemsep0em
\item A novel, privacy-first approach to the training of data-driven text anonymization models in the absence of labeled data.
\item An implementation of that approach with a large knowledge graph derived from Wikidata, which is applied to automatically label biographical texts from Wikipedia with text spans to mask.
\item A new dataset\footnote{The dataset along with the guidelines is publically available: \url{https://github.com/anthipapa/textanonymization}} of 553 Wikipedia summaries manually annotated for sensitive information, which we use to evaluate the empirical performance of the proposed approach.
\end{itemize}

The rest of the paper is structured as follows. The next section reviews related work on text anonymization. Section \ref{sec:approach} describes the three steps of our approach, which is then evaluated in Section \ref{eval}. We conclude in Section \ref{sec:conclusion}.

\section{Related Work}

As stipulated by Article 8 of the European Convention
on Human Rights and Article 12 of the Universal Declaration of Human Rights, privacy is a fundamental right and an essential component of a democratic society. To ensure that every person remains in control over their own personal data, legal frameworks such as the General Data Protection Regulation (GDPR) \cite{gdpr} in the European Union spell out how personal data should be collected, processed and shared. 

Personally identifiable information can be divided into main categories \cite{soton399692}:

\begin{description}\label{identifiers}
\itemsep0em
    \item \textbf{Direct identifiers:} information that can be used to directly single out an individual, such as the person name, social security  or passport number, email address, bio-metric records, mobile phone number, etc.
    \item \textbf{Quasi identifiers:} information that is not univocally related to a unique individual, but may lead to identification when combined with other quasi identifiers\footnote{For instance, the combination of gender, birth date and postal code has been shown to single out between 63 and 87\% of the U.S. population \cite{golle2006}.} such as date of birth, gender, ethnicity, religion, etc.
\end{description}

Although most existing work on anonymization focuses on quantitative tabular data, several studies have also investigated the problem of anonymizing text data, either from an NLP perspective or from the field of data privacy, and in particular privacy-preserving data publishing (PPDP).

 The first NLP approaches relied on rule-based or rule-assisted methods for pattern detection mainly in medical text documents \cite{ruch_medical_2000,1588104}. Recent approaches on the anonymization of medical health records focus on detecting direct identifiers and quasi identifiers using sequence labelling models trained from manually annotated data, focusing on pre-defined categories. \cite{10.1136/amiajnl-2012-001012,dernoncourt2017identification,10.1016/j.jbi.2017.05.023,wikipii}. 
 
 One drawback of these NLP approaches is that they are typically limited to detecting predefined (semantic) categories of identifiers and quasi identifiers, without taking into account other types of information that may uncover the identity of the person. For instance, the physical appearance of a person or their professional activities will often provide clues about the person identity, yet rarely belong to the semantic categories to detect.  In addition, those methods typically mask all detected text spans uniformly, without making it possible to parameterize the anonymization process based on the estimated disclosure risk. 
 
 PPDP approaches to text anonymization, on the other hand, generally seek to enforce a \textit{privacy model} such as $k$-anonymity~\cite{Samarati98}, then search for the optimal set of masking operations -- such as removal or generalization of the original values -- to ensure the privacy model requirements are met. 
 
 The $k$-anonymity model was adapted for text data in the $k$-safety and $k$-confusability models \cite{chakaravarthy2008efficient,Cumby2011AML}. Both approaches require every sensitive entity to be indistinguishable from at least $k-1$ other entities. The entities are then generalized to become indistinguishable and thus, safe from disclosure risk. The $t$-plausibility model \cite{anandan} introduced a similar approach based on the generalization of (already detected) terms, seeking to ensure that at least $t$ documents can be derived through specialization of the generalized terms. A final model is $C$-sanitize \cite{Sanchez2016} which provides a priori privacy guarantees by relying on the mutual information of the sensitive entities and the rest of the words in the document. The words that are more likely to lead to identification of the sensitive term to be protected, either individually or in combination with others, are then generalized. The mutual information scores used in $C$-sanitize are derived from co-occurrence counts in web data.
 
 PPDP approaches makes it possible to explicitly adjust the trade-off between data protection and data utility. However, many PPDP approaches rely on the assumption that sensitive entities are already detected in a preprocessing step. They also often rely on external resources that may be difficult to gather, such as exhaustive collections of contexts for all entities to protect \cite{acl2021}.
 
 Finally, recent work has investigated the use of differential privacy \cite{10.1561/0400000042} to generate synthetic texts \cite{9415242} or obfuscate documents to protect them against authorship attribution \cite{DBLP:conf/post/FernandesDM19,feyisetan2019leveraging}. However, those methods operate by introducing artificial noise either in the text or in the word representations derived from it. Contrary to the NLP and PPDP methods detailed above, those methods do not preserve the ``truth value'' of the document, and seek therefore to address a slightly different task than text anonymization. This aspect is particularly important for use cases related to clinical or legal NLP, where the anonymization process is typically not allowed to alter the document content beyond masking or generalising text spans containing personal information. For instance, a clinical report in which the description of symptoms, diagnostic, chosen treatment and clinical outcomes is not guaranteed to stem from a real patient would be of little interest to medical professionals. The same holds for court rulings in which the wording employed by the court (with the exception of personal identifiers) must generally be preserved without tampering.

\section{Approach}
\label{sec:approach}
 In the following subsections, we present the three main components of our approach. 
 
 \subsection{Step 1: Modeling of background information} \label{step1}

The term \textit{background information} refers to an attacker's possible additional knowledge that could be used to re-identify an individual. Depending on the type of data to be anonymized, this information, as well how it was acquired by the attackers can vary. It is, of course, difficult to define exactly what this knowledge might be, but general assumptions can be useful \cite{desan}.

A convenient way to express this background information is through a knowledge graph connecting individuals with their various personal identifiers. This knowledge graph can be extracted from a variety of sources, such as structured databases, social network data or co-occurrence counts on web data \cite{Sanchez2016}.

% , and general sources can even be used for more specialised fields like the medical field \cite{feyisetan2019leveraging}.

However, knowledge graphs do not provide any efficient mechanism for determining the number of individuals associated with a particular combination of \mbox{(quasi-)identifiers}. This is particular problematic for quasi-identifiers that may be shared by a large set of individuals (e.g. male or female). To this end, we construct an \textit{inverted index}\footnote{ An inverted index is a data structure commonly used in information retrieval, and consists of an index mapping terms to the documents they occurred in \cite{ir}. Those documents are represented through a sorted list of indices, making it possible to efficiently compute intersections.} from the knowledge graph. In our case, the inverted index associates terms to unique indices of each individual associated with this term. Figure \ref{fig:intro} includes an example of inverted index where the individual with index=4 is connected with the terms ``Leandro Micael Gomes Albano'', ``Lisbon'' and ``Freamunde''.

Based on this inverted index, one can then efficiently query the data structure to determine the list of individuals that are related to a given set of terms. This query can be implemented through a Boolean retrieval model, taking advantage of the fact that the postings are already sorted to compute their intersection. If the resulting set is a singleton, this means that the combination of terms allows us to uniquely re-identify the person. This is for instance the case for the combination of terms ``Lisbon'' and ``Freamunde'' in Figure \ref{fig:intro}.

One important benefit of using an inverted index to capture the relation between individuals and their quasi-identifiers is the fact that it can be easily extended to incorporate variations of identifiers. For instance, dates and person names can be expressed in multiple formats, common nouns may have synonyms, and even locations may have alternative written variants, such as Lisbon vs. Lisboa.

\subsection{Step 2: Text Anonymization with Distant Supervision}\label{step2}

Using documents related to individuals present in this knowledge graph, we can then automatically determine which terms to mask through queries on the inverted index. The first step is to search for term occurrences in the text using a gazetteer, as illustrated in Figure \ref{fig:intro}. 

Only some of the terms located by the gazetteer will need to be masked. We rely on the $k$-anonymity privacy model to account for the disclosure risk associated with a given set of terms in a document. $k$-anonymity was first introduced by \newcite{Samarati98} and requires every sensitive entity to be indistinguishable from at least $k-1$ other entities based on their attributes. Through $k$-anonymity, the individuals can be 'hidden' by being part of a larger group. The value of \textit{k} can vary depending on the dataset that needs protecting, but it should be larger than 1, since \textit{k=1} means no anonymity. A common recommendation is to use \textit{k=5} \cite{10.1197/jamia.M2716}, which we follow in our experiments.

\renewcommand{\listingscaption}{Algorithm}
\begin{listing}[t!]
\begin{minted}[escapeinside=||,mathescape=true,numbers=right,fontsize=\scriptsize, fontfamily=sf, frame = single]{python}

def getTermsToMask(terms, postings, maxArity, 
       termSelect, k):
  # terms: set of terms found in a document
  # postings: inverted index
  # maxArity: maximum arity of the term combinations
  # termSelect: greedySelect or randomSelect (see below)
  # k: k-anonymity value to satisfy 
    
  termsToMask = |$\emptyset$|
  
  # We mask terms associated with $<k$ individuals
  for term in terms:
    if len(postings[term]) |$< k$|:
      Add term to termsToMask
  
  while True:

    # We create a set of possible term combinations, 
    # starting with pairs, then triples, etc. 
    termsTuples |$\leftarrow \emptyset$|
    for arity in [2,...maxArity]:
      newTuples |$\leftarrow$| combine(terms - termsToMask, arity)
      termTuples |$\leftarrow$| termTuples + newTuples
      
    # For each term combination, we check whether the 
    # intersection of postings gives $<k$ individuals
    for (term|$_1$|,...term|$_n$|) in termTuples:
    
      if |$1 \leq$| len(|$\bigcap_{i=1}^{n}$| postings[term|$_i$|]) |$< k$|:
      
        # If yes, we select a term to mask
        selectedTerm |$\leftarrow$| termSelect(term|$_1$|,..., term|$_n$|, postings)
        Add selectedTerm to termsToMask
        # and restart the evaluation of term combinations
        break
    
    else: # stop when all combinations satisfy k-anonymity
      break
    if terms = termsToMask:  # or if all terms are masked
      break

  return termsToMask
  
def greedySelect(terms, postings):
  # greedy selection: select term with shortest posting list
  return |$\argmin_{\text{term}_i \in \text{terms}} $| postings[term|$_i$|] 
  
def randomSelect(terms, postings):
  return select random term from terms

\end{minted}
\caption{Extraction of terms to mask in a document, based on $k$-anonymity and posting lists mapping each possible term to the list of persons associated with it. When a combination of quasi-identifiers breaks the $k$-anonymity constraint, we either select the term with the shortest posting (\begin{footnotesize}\textsf{greedySelect}\end{footnotesize}), or choose a random term (\begin{footnotesize}\textsf{randomSelect}\end{footnotesize}).}
\label{algo:anonymity}
\end{listing}

Algorithm \ref{algo:anonymity} is employed to determine the terms to mask in a document based on the posting lists. The algorithm starts (lines 11-14) by checking whether some terms need to be directly masked (as their presence would break $k$-anonymity). This is for instance the case for the term ``Leandro Micael Gomes Albano'', which is related to a single individual. The procedure continues by forming gradually more complex combinations of terms, and computing the intersection of their posting lists (lines 27-29).  Intersections of size $< k$ represent a breach of $k$-anonymity, and imply that at least one of their terms must be masked. Several strategies can be followed to determine which term is most useful to mask in each combination. In this work, two strategies have been implemented. The greedy strategy (lines 45-47) consists in systematically masking the most specific term -- that is, the term with the shorted posting list. Alternatively, one can also select at random the term to mask in each combination.

\subsection{Step 3: Fine-tuning}\label{step3}

The two steps above result in an automatically annotated dataset that can be directly used to fine-tune a language model. Thisincreases the ability of the model to generalize to texts and individuals not covered in the knowledge graph.

We frame the problem of text anonymization as a token-level sequence classification task. In this paper, we rely on BERT \cite{devlin2019bert}, a large, transformer-based language model employed in many sequence classification tasks in the field of NLP, including recent work on data privacy \cite{alsentzer-etal-2019-publicly}. More specifically we use a RoBERTa model \cite{roberta} for our experiments. As in most distant/weak supervision frameworks \cite{mintz-etal-2009-distant,Ratner:2017:SRT:3173074.3173077}, the training of a generic, neural model allows us to process arbitrary texts without depending on the availability of external resources such as knowledge graphs. 

\section{Evaluation} \label{eval}

The proposed approach is evaluated on short biographical texts extracted from Wikipedia, using graph data from Wikidata to determine the terms to mask to ensure $k-$anonymity. We first present the document collection and knowledge graph, and then describe a manually annotated test set of biographies employed to assess the performance of the fine-tuned RoBERTa models. We then present our results and discuss them.

\subsection{Distant labelling of Wikipedia articles}

The relevant background knowledge for this task comes from Wikidata\footnote{\url{https://www.wikidata.org/}}. Wikidata provides structured data, acquired and maintained collaboratively. It is at times used directly by Wikipedia, but typically restricted to the creation of the page's infobox. We assume that this is all the possible knowledge an attacker could acquire and use against our dataset. At this stage this knowledge graph is limited, but can be expanded to include more information or altered to be domain-specific.

The knowledge graph employed in this work consists of entities such as names, nicknames, translations, professions, places of birth and death, and more. To handle entities that may have several surface realizations, we augmented the inverted index to include all possible variants of a given term. This includes dates (e.g. 1992-08-05 $\rightarrow$ 5 May, 1992),  person names (``Leandro Albano'' $\rightarrow$ ``L. Albano''), country-nationality pairs (Austrian-Austria), and alternative names for locations. A white list of very frequent terms was also established to filter out common words from the knowledge graph that are deemed generic enough not to necessitate any masking, as for example "born", "age", "man", "woman". The resulting inverted index comprises 22 034 977 terms.

This knowledge graph was then applied on a dataset of short Wikipedia biographies \cite{wikidataset} whose entries were filtered to consist of only humans that are also present in the knowledge graph, resulting in a total of 502 678 distinct biographies. The dataset was already split into training (80\%), validation (10\%), and test datasets (10\%), which was preserved in this evaluation. The biographical texts are about 4 sentences long on average, with a standard deviation of 3.58.

\subsection{Evaluation data}
\label{sec:annotation}

We conduct a manual annotation effort on a subset of summaries for evaluation purposes. A random sample of 553 summaries was extracted from the test dataset. The distribution of summary lengths reflects that of the test dataset, with the average being 4 sentences (11\%), while around 65\% were summaries with less than the average. The largest summary in the sample was 20 sentences long.

For the manual annotation, the TagTog \footnote{\url{https://www.tagtog.net/}} tool was used with 5 annotators, four undergraduate students in law, and one NLP researcher. These annotators were already familiar with the annotation task, as they had been trained and conducted similar annotation efforts in the past. They were also provided with detailed annotation guidelines and examples to follow. The objective of the annotation was to (1) find terms associated with personal information and (2) decide which of those terms ought to be masked to conceal the identity of the individual in the biography.

Of the 553 summaries, 20 biographies were annotated by two annotators, and the rest by a single annotator. The annotators were provided with pre-annotations to mark terms that were likely to express personal information. Those pre-annotations were generated by combining the gazetteer (see Section \ref{sec:approach}) with a neural NER model and a set of heuristics to recognize dates and numerical values. It should, however, be stressed that the annotators were explicitly instructed to only use those pre-annotations as a starting point and correct them as they see fit -- either by modifying/deleting terms that did not include any personal information, or by inserting new terms that were ignored by the pre-annotations. See the Appendix for two annotation examples.

After modifying the spans, the annotators were also tasked with correcting or assigning a semantic category to the span. In the pre-annotations this label was given by the NER model, while MISC was used for entities taken from the inverted index and did not fall into any of the model's types. The categories along with some examples and basic statistics are shown in Table \ref{tbl:sem}.

\begin{table*}[t!]
\begin{center}
\resizebox{\textwidth}{!}{%
\begin{tabular}{llll}
\toprule
 Entity type & Examples & \# of mentions & \% masked \\
  \midrule
  PERSON & names, nicknames, spelling variations - translations & 2005 (17\%) &  99\% \\
  LOC & cities, countries, infrastructures & 979 (8.7\%) &  75\% \\
  ORG & schools, universities, churches & 1667 (14.8\%) &  84\% \\
  DEM & nationalities, job titles, education, health information & 2180 (19.4\%) & 27\%  \\
  DATETIME & dates, time, durations & 2158 (19.2\%) &  84\% \\
  QUANTITY & percentages, meters, monetary values & 403 (3.5\%) & 74\%  \\
  MISC & miscellaneous (not part of above categories) & 1806 (16\%) &  57\% \\
\end{tabular}
}
\caption{\label{tbl:sem} Statistics on the semantic type.}
\end{center}
\end{table*}

% \begin{description}
% \item PERSON (e.g. names, nicknames, spelling variations, name translations)
% \item LOC (e.g. cities, countries, infrastructures) 
% \item ORG (e.g. schools, universities, churches)
% \item DEM (e.g. nationality, job title, education, health related information)
% \item DATETIME (e.g. dates, time, time and duration)
% \item QUANTITY (e.g. percentages, meters, monetary values)
% \item MISC (e.g. any type of information about an individual that cannot fit into the previously mentioned categories)
% \end{description}
% \raggedbottom

After this initial step of term detection, the annotators had to determine which of these terms could lead to the identification of the individual, either as direct or quasi-identifiers (see Section \ref{identifiers}).  Each term is therefore labeled as one of three mutually exclusive identifier types: \begin{description}
\itemsep0em
    \item[DIRECT] if the term denotes a direct identifier
    \item[QUASI] if the term denotes a quasi identifier
    \item[NO\_MASK] if the term can be left in clear text without impairing $k-$anonymity
\end{description}

Table \ref{tbl:idtype} shows the number of instances per identifier type.

% \begin{table*}[t]
% \begin{center}
% \begin{tabular}{ccc}
% \toprule
%  Entity type & \# of instances & \% \\
%   \midrule
%   DIRECT & 1579 & 14\% \\
%   QUASI & 6281 & 56\% \\
%   NO\_MASK & 3357 & 30\% \\
% \end{tabular}
% \caption{\label{tbl:idtype} Statistics related to identifier type}
% \end{center}
% \end{table*}

% \begin{table*}[t!]
% \begin{center}
% \begin{tabular}{ccc}
%     \toprule
%     %    Labels &
%         Level &  Kappa & Alpha \\   
%         \midrule
% %    \midrule
% %	 \multirow{2}{*}{Semantic type} & Span && 0.64 && 0.84 \\
% %	 & Character && 0.85 && 0.87  \\
% 	 Span & 0.44 & 0.59 \\
% 	 Character & 0.81 &  0.73 \\
% % 	 \multirow{2}{*}{Identifier type} & Span  & 0.71 & 0.46 & 0.51 & 0.68 & 0.67 \\
% % 	  & Character & 0.94 & --  & -- & 0.69 & --\\
% \end{tabular}
% \caption{\label{tbl:iaa} Inter-annotator agreement on the identifier type (DIRECT, QUASI OR NO\_MASK).}
% \end{center}
% \end{table*}

\begin{table*}[t!]
    \begin{minipage}{.5\textwidth}
      \centering
\begin{tabular}{ccc}
\toprule
 Entity type & \# of instances & \% \\
  \midrule
  DIRECT & 1579 & 14\% \\
  QUASI & 6281 & 56\% \\
  NO\_MASK & 3357 & 30\% \\
\end{tabular}
\caption{\label{tbl:idtype} Statistics on the identifier type.}
    \end{minipage}%
    \begin{minipage}{.5\textwidth}
      \centering
        \begin{tabular}{ccc}
    \toprule
    %    Labels &
        Level &  Kappa & Alpha \\   
        \midrule
%    \midrule
%	 \multirow{2}{*}{Semantic type} & Span && 0.64 && 0.84 \\
%	 & Character && 0.85 && 0.87  \\
	 Span & 0.44 & 0.59 \\
	 Character & 0.81 &  0.73 \\
%	 &&& 
% 	 \multirow{2}{*}{Identifier type} & Span  & 0.71 & 0.46 & 0.51 & 0.68 & 0.67 \\
% 	  & Character & 0.94 & --  & -- & 0.69 & --\\
\end{tabular}

\caption{\label{tbl:iaa} Inter-annotator agreement on the identifier type.}
    \end{minipage} 
\end{table*}

% Regarding semantic type, the most prominent categories were DEM (19\%), DATETIME (19\%), PERSON (20\%), MISC(16\%) and ORG (14\%), while regarding identifier type 56\% of the masked tokens were quasi identifiers, 30\% were left unmasked, and 14\% were direct identifiers.

For the 20 multi-annotated texts, we calculated inter-annotator agreement on the identifier type, by calculating Cohen's $\kappa$, as well as Krippendorff's $\alpha$ on the span and on the character level \cite{agree}, with the first being based on agreement and the latter based on disagreement. The results are summarized in Table \ref{tbl:iaa}. Inter-annotator agreement for identifier types is not a reliable measure of the quality of the annotations in this setting, since there may be several equally correct solutions to a given anonymization task \cite{acl2021}, which is also reinforced by the disagreement between annotators. Out of the 74 disagreements, 64 were between the QUASI and NO\_MASK label pairs, while 10 were between the DIRECT and QUASI label pairs.

\subsection{Distant supervision models} \label{sec:models}

We use the automatic annotations from the greedy and the random functions to train two RoBERTa models with a linear inference layer on top (\textit{GreedyBERT, RandomBERT}). The parameters used to train the models can be found in Table \ref{tbl:Params} in the appendix.

% We used an IOB scheme to account for multi-token annotations, so that each token received either a B-MASK, I-MASK or an 'O' label.

We evaluate the performance of the models both against the automatically labeled development and test data, and on the manually annotated dataset of 553 biographies. We calculate precision, recall, and $F_1$-score on the entity level, modifying our script to account for cases  when there is a partial match between the prediction and the true string regarding boundaries, but where the latter is always masked (e.g. "from Winnipeg", instead of "Winnipeg").

\subsection{Experimental Results}

The evaluation on the automatically annotated data is meant to evaluate the feasibility of the approach, i.e., to test whether or not the background knowledge allows for a learnable annotation, whereas the manually annotated data is employed to assess the generalizability of the approach. Table \ref{tbl:perf2} shows the result for the first type of evaluation, which shows comparable performance.

\begin{table*}[t!]
\begin{center}
\begin{tabular}{llll|lll}
\toprule
 & & Dev & & & Test\\
  \midrule
 & Precision & Recall & F1 score & Precision & Recall & F1 score \\
%    \midrule
  GreedyBERT & 0.895 &  0.911 & 0.903 &  0.883 &  0.910 &  0.897 \\
\midrule
RandomBERT & 0.811 & 0.928 & 0.866 & 0.801 & 0.921 &  0.860 \\ \\
\end{tabular}
\caption{\label{tbl:perf2} Entity-level precision, recall and $F_1$ scores of greedy and random RoBERTa on the automatically labeled data. The scores account for exact matches, as well as modulo postprocessing to ignore boundary mismatches due to punctuation or prepositions}
\end{center}
\end{table*}

We then test the performance of the models on the manually annotated dataset. To contrast the annotation provided by our approach with a standard named-entity annotation task, we also run a named entity recogniser based on the RoBERTa language model \cite{roberta} and fine-tuned for NER on the Ontonotes v5 \cite{ontonotes2011}, as implemented in spaCy\footnote{\url{https://spacy.io}}. Since the manually annotated dataset also included information on the identifier type of each masked term, we also calculate recall for direct and quasi identifiers separately. Recall is the most critical metric for anonymization tasks since false negatives could directly lead to re-identification. These results are shown in Table \ref{tbl:broken}.

% \begin{table}[t!]
% \begin{center}
% \begin{tabular}{*3c}
% \toprule
%  & Recall direct & Recall quasi \\
%  & identifiers & identifiers \\
%  \midrule
% GreedyBERT   & 0.827 & 0.613 \\
% RandomBERT   & 0.811 &   0.652  \\
% \midrule
%  Neural NER   & 0.810  & 0.801   \\
% \end{tabular}
% \caption{\label{tbl:broken} Entity-level recall on direct and quasi identifiers for the manually annotated dataset}
% \end{center}
% \end{table}

\begin{table*}[t!]
\begin{center}
\begin{tabular}{*6c}
\toprule
 & Precision & Recall$_{all}$ & Recall$_{direct}$& Recall$_{quasi}$ & F1 score\\
 \midrule
 RoBERTa NER & 0.770 & 0.845 & 0.810 & 0.801 & 0.805\\
 \midrule
GreedyBERT   & 0.669 & 0.836 & 0.898 & 0.774 & 0.743 \\
RandomBERT  & 0.650 & 0.832 & 0.895 & 0.770 &  0.730 \\
\end{tabular}
\caption{\label{tbl:broken} Entity-level precision, recall and $F_1$ score of greedy and random RoBERTa on the manually labeled dataset of 553 Wikipedia biographies, compared to the results obtained by a RoBERTa neural language model fine-tuned for Named Entity Recognition on Ontonotes v5.}
\end{center}
\end{table*}

We also assess the empirical performance of the greedy and random models on the recently released TAB corpus \cite{tab}, which comprises 1,268 court cases from the European Court of Human Rights, annotated to evaluate text anonymization methods. This allows us to determine the robustness of the proposed distant supervision approach across different domains. The cases were annotated following similar guidelines to the manually annotated Wikipedia biographies. Table \ref{tbl:tab_} shows the results on the test set of the TAB corpus. The Table also includes the performance of the neural NER model as well as a Longformer model trained on the training set of the TAB corpus, as detailed in \cite{tab}.

% We also calculate the paper's proposed weighted precision score, which uses a BERT language model to assign a weight to the tokens depending on how informative they are. This means that more specific tokens will be assigned a higher weight while calculating the precision score. 

%  that measure the degree of data protection of the system

% \begin{table*}[t!]
% \begin{center}
% \begin{tabular}{*7c}
% \toprule
%  &  Recall$_{all}$ & Recall$_{direct}$ & Recall$_{quasi}$ & F1 score & P$_{all}$ & WP$_{all}$\\
%  \midrule
% GreedyBERT   & 0.814 & 0.782  & 0.847 & 0.394 & 0.260 & 0.291\\
% RandomBERT  & 0.668 & 0.530 & 0.806 & 0.377 & 0.263 & 0.292\\
% \midrule
%  Neural NER  & 0.906  &  0.940 & 0.874 & 0.565 & 0.441 & 0.515\\
% \end{tabular}
% \caption{\label{tbl:tab_}  Total and entity-level recall, f1 score, as well as precision and weighted precision of greedy and random RoBERTa models on the TAB test dataset. Included are the results from the neural NER model (pre-trained RoBERTa language model fine-tuned on Ontonotes v5)}
% \end{center}
% \end{table*}

\begin{table*}[t!]
\begin{center}
\begin{tabular}{*6c}
\toprule
 & Precision &  Recall$_{all}$ & Recall$_{direct}$ & Recall$_{quasi}$ & F1 score  \\
 \midrule
RoBERTa NER  & 0.441 & 0.906  &  0.940 & 0.874 & 0.565 \\
Longformer fined-tuned on TAB & 0.836 & 0.919  & 1.000 & 0.916 & 0.876 \\
 \midrule
GreedyBERT  & 0.260 & 0.814 & 0.782  & 0.847 & 0.394  \\
RandomBERT  & 0.263 & 0.668 & 0.530 & 0.806 & 0.377  \\
\midrule
\end{tabular}
\caption{\label{tbl:tab_}  Entity-level precision, recall and $F_1$ score on the test set of the Text Anonymization Benchmark (TAB). The tables compares the greedy and random RoBERTa against the RoBERTa model fine-tuned on Ontonotes and a Longformer model fine-tuned on the training set of TAB.}
\end{center}
\end{table*}

\subsection{Discussion}

The performance of a model trained using distant supervision will necessarily depend on the quality and coverage of the knowledge graph employed to generate the labels. In this work the coverage of the knowledge graph (and of the inverted index derived from it) will influence (1) which terms will be considered as personal information and (2) which of those terms will need to be masked to enforce $k$-anonymity. 

The experimental results illustrate some of the limitations of using Wikidata as background knowledge. There were many instances of information mismatch between Wikipedia and Wikidata (e.g. different name spellings, information present in Wikipedia but not Wikidata). This led to either some PII not being part of the annotations or being partially annotated, which also resulted in the models often deciding to mask parts of entities instead of the entire spans. On the other hand, the automated masking based on the inverted index also led to some spurious masking decisions, notably for rare terms that do not express PII but are tied to a small set of individuals in Wikidata.

When testing the models on the manually annotated Wikipedia (Table \ref{tbl:broken}), we see that the performance of the two RoBERTa models is better for direct identifiers compared to the the neural NER system, with the latter outperforming the models on the rest of the metrics. The lower precision of the models is an indication of the aforementioned issue of background information choice, since the models tend to mask information that would not generally be considered a PII, due to the presence of similar terms in the inverted index, or mask terms that the annotators decided not to mask, while the low recall can also be attributed to the models' masking decisions not being a part of the annotations.

While also analyzing the masking decisions made by the annotators we observe an over-masking trend, as well as a tendency to mask named entities to a larger extent, especially for longer texts since those are the 'safer' choices (e.g. as shown in Table \ref{tbl:sem}, the most prominently masked categories were PERSON, ORG, and DATETIME while regarding identifier type 56\% of the masked tokens were quasi identifiers, and only 30\% were left unmasked, as shown in Table \ref{tbl:idtype}). As mentioned above, there is no gold answer regarding the set of masking decisions, as long as the identity of the individual is protected, as also shown by the low recall on quasi identifiers in Table \ref{tbl:broken}. For this reason we manually compared the output of the models against both the neural NER system and the manual annotations for a few texts, and we observed that despite their low scores, the two RoBERTa models often offer masking decisions, which despite not being similar to that of the annotator(s) or complete in the sense of entity boundaries, were able to prevent identification. 

% \begin{description}
% \itemsep0em
% \item \textbf{Original Text}\\
% \begin{small}Jenn Mierau is a Canadian electropop musician originally from Winnipeg, who is now based in Montreal.\end{small}
% \item \textbf{Human annotator}\\
% \begin{small}******* is a Canadian electropop musician originally from *******, who is now based in Montreal.\end{small}
% \item \textbf{Mask from supervised NER model}\\
% \begin{small}******* is a ******* electropop musician originally from *******, who is now based in *******.\end{small} 
% \item \textbf{Mask from distantly supervised BERT model}\\
% \begin{small}******* ******* is a Canadian ******* originally *******, who is now based in Montreal.\end{small}
% \end{description} 
% \raggedbottom
% \noindent\rule{0.4\textwidth}{0.2pt}
\begin{description}
\itemsep0em
\item \textbf{Original Text}\\
\begin{small}Jenn Mierau is a Canadian electropop musician originally from Winnipeg, who is now based in Montreal.
\end{small}
\item \textbf{Human annotator}\\
\begin{small}******* is a Canadian electropop musician originally from *******, who is now based in Montreal.
\end{small}
\item \textbf{Mask from supervised NER model}\\
\begin{small}******* is a ******* electropop musician originally from *******, who is now based in *******.
\end{small}
\item \textbf{Mask from distantly supervised RoBERTa model}\\
\begin{small}******* ******* is a Canadian ******* musician originally *******, who is now based in Montreal.
\end{small}
\end{description}
\raggedbottom

The distantly supervised model produces mask a set of decisions that include direct identifiers (full name), but also over-masks (e.g. "from Winnipeg" instead of "Winnipeg") or masking spans that the annotators did not choose to mask (e.g. musician). Despite the model's masking decisions preventing identification, this behavior is not reflected in the evaluation results. The combination of "Canadian", "musician" and "Montreal" does not lead to re-identification (excluding Wiki-related pages since this is the source of the data).

Partial masking of spans as reflected in our evaluation only covers specific cases when the boundaries of the prediction are slightly different than those of the annotated span (e.g. masking "from Winnipeg" instead of just "Winnipeg" is a correct masking decision). More importantly, the reason behind the performance of the model is the lack of representation of the model's predictions in the pool of acceptable answers, as annotated by human annotators. This is reflected in examples like "musician" being masked.

By evaluating the performance of the models on domain specific data, we observe a lower recall on direct identifiers (Table \ref{tbl:tab_}) compared to that on the Wikipedia dataset. This is due to direct identifiers that were not present in the training data (e.g. codes). The higher recall score on quasi identifiers along with the very low precision score, shows that the models over-masked the text, resulting in very low data utility.
% The table also shows a slight difference between the precision and the weighted precision scores, indicating that the words masked have a low level of informativeness. 

Court cases contain a lot of information that is not directly relevant to the case (e.g. organisation names or locations) which should not be masked. The two RoBERTa models and the NER model mask a large number of spans, regardless of the level of informativeness, based on their low precision score. This trend shows that all generic systems tend to over-mask text that is domain specific, especially compared to a model directly trained on the manual annotations, as shown in Table \ref{tbl:tab_}, which shows both high recall and high precision scores meaning that it strikes a balance between data utility and privacy risk.

\section{Conclusion}
\label{sec:conclusion}

We proposed a novel method to automatically annotate text documents containing personal information using background information expressed as a knowledge graph. The long-term objective of such an approach is to bootstrap text anonymization models in the absence of supervised training data, using distant supervision to determine which text spans to mask to enforce a privacy model such as $k$-anonymity. The automatically annotated documents can then be employed to fine-tune a pretrained language model. 

An implementation of the approach using Wikipedia biographies and Wikidata as background information is presented. We evaluate the approach on a manually annotated set of biographies and domain specific data.  Our experimental results demonstrate that the performance of the approach is heavily dependent on the choice of background knowledge. The results, especially when compared to actual model output, illustrate the challenge of evaluating such a task when the acceptable pool of possible masking solutions is not limited to just one answer, as well as the need for an extensive and broad knowledge graph.

Future work will investigate several research directions. One important issue is enhancing the quality of the knowledge graph to improve the coverage of quasi-identifiers, while filtering out spurious terms that do not express PII. Furthermore, we aim to extend the inverted index with other sources of background knowledge beyond structured databases, and in particular co-occurrence estimates from raw, web-scale data. Such broad knowledge graphs can then be adapted and used for another given domain.

\section{Acknowledgements}

We acknowledge support from the Norwegian Research Council (CLEANUP project6, grant nr.308904).

\section*{Appendix}
\label{sec:appendix}

\begin{figure*}[t]
% \hspace*{-1cm} 
\includegraphics[width=0.99\textwidth]{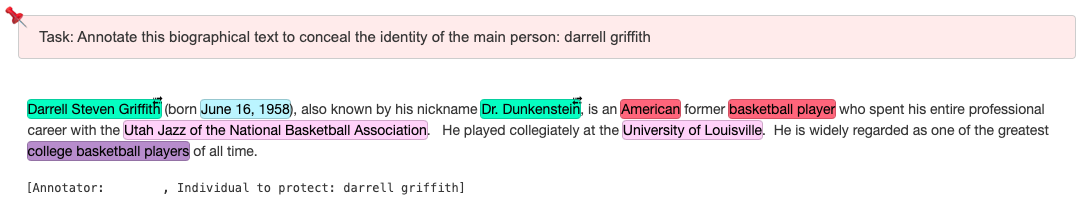}
  \caption{Step 1 of the annotation process: identification of entities that express personal information.}
  \label{fig:tagtog1}
\end{figure*}

\begin{figure*}[t]
% \hspace*{-1cm} 
\includegraphics[width=0.99\textwidth]{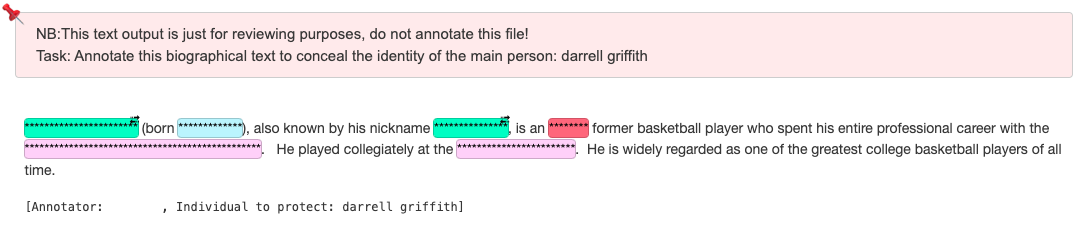}
  \caption{Step 2 of the annotation process: decision on which entities to mask in the text.}
  \label{fig:tagtog2}
\end{figure*}

Example of the setup for the annotation task mentioned in Section \ref{sec:annotation}. Figure \ref{fig:tagtog1} shows the initial entities that were identified as expressing personal information (\textit{Step 1}), and Figure \ref{fig:tagtog2} illustrates the final result with the entities that the annotators decided to mask replaced with *** (\textit{Step 2}). 

Table \ref{tbl:Params} shows the parameters used to train greedy and random RoBERTa on the automatically annotated datasets, mentioned in Section \ref{sec:models}.

\begin{table}[h]
 %\vspace{-2mm}
\begin{center} 
\begin{tabular}{lcc}  
	\toprule  %\vspace{-1mm}
	 Parameter &  \\
	\midrule 
     Optimizer & AdamW  \\ 
     Learning rate & 2e-5\\ 
     Loss function & CrossEntropy \\
     Loss function MASK-Weight & 10 \\
     Inference layer & Linear \\
     Epochs & 2 \\
     Full fine-tuning & yes \\
     GPU & yes \\
     Early stopping & yes \\
	%\bottomrule
\end{tabular}
\caption{\label{tbl:Params} Training Parameters for RoBERTa models}
\end{center}
\end{table}

% \nocite{*}
\section{Bibliographical References}\label{reference}
%\label{main:ref}

\bibliographystyle{lrec2022-bib}
\bibliography{lrec2022-example}

\end{document}